\documentclass{article}
\usepackage[utf8]{inputenc}
\usepackage{authblk}
\usepackage{setspace}
\usepackage[margin=1.25in]{geometry}
\usepackage{graphicx}
\graphicspath{ {./figures/} }
\usepackage{subcaption}
\usepackage{amsmath}

\usepackage{booktabs}
\usepackage{graphicx}
\usepackage{algorithm}
\usepackage{algpseudocode}
\usepackage{amsmath}
\usepackage{fancyhdr}
\usepackage{setspace}
\usepackage{svg}
\usepackage[justification=centering]{caption}

\usepackage[
  backend = biber,
  style = authoryear
]{biblatex}

\DeclareUnicodeCharacter{0301}{\'{e}}
\title{Lexico-semantic and affective modelling of Spanish poetry: A semi-supervised learning approach}

\author[1,2,*]{Alberto Barbado}
\author[3,4]{Mar\'{i}a Dolores Gonz\'{a}lez}
\author[ ]{D\'{e}bora Carrera}

\affil[1]{Universidad Polit\'{e}cnica de Madrid, Departamento de Inteligencia Artificial, Madrid, Spain}
\affil[2]{Telef\'{o}nica, Madrid, Spain}
\affil[3]{IFEMA, Madrid, Spain.}
\affil[4]{Universidad Oberta De Catalunya, Barcelona, Spain}
\affil[*]{Corresponding author email: alberto.barbadogonzalez@telefonica.com}

\date{}

\begin{document}

\maketitle

\begin{abstract}
Text classification tasks have improved substantially during the last years by the usage of transformers. However, the majority of researches focus on prose texts, with poetry receiving less attention, specially for Spanish language. 
In this paper, we propose a semi-supervised learning approach for inferring 21 psychological categories evoked by a corpus of 4572 sonnets, along with 10 affective and lexico-semantic multiclass ones. The subset of poems used for training an evaluation includes 270 sonnets. With our approach, we achieve an AUC beyond 0.7 for 76\% of the psychological categories, and an AUC over 0.65 for 60\% on the multiclass ones. 
The sonnets are modelled using transformers, through sentence embeddings, along with lexico-semantic and affective features, obtained by using external lexicons. Consequently, we see that this approach provides an AUC increase of up to 0.12, as opposed to using transformers alone.
\end{abstract}


\section{Introduction}

\noindent Text mining techniques are useful for discovering patterns and extracting insights from within different sources of text data sets. This is useful for several tasks, such as classifying texts into opinions, sentiments, emotional states or topics, by modelling the input texts through its lexico-semantic, affective or semantic meaning \parencite{liu2012sentiment}.

Most of the research studies focus on prose texts \parencite{berry2004survey}, with less analyses for poetic documents, mainly because of the complexity of its meaning (like metaphors), and structure (like rhyme or enjambment) \parencite{jakobson1960linguistics}. Another problem is that there is a lack of poetry corpora \parencite{kaplan2007computational}. Nonetheless, poetry computational analysis is growing within the literature, and studies have shown that several text mining task can be tacked through similar approaches used for prose texts \parencite{greene2010automatic, kao2012computational}.

Following this, poetry classification can be achieved through Machine Learning (ML), either from a supervised approach \parencite{lou2015multilabel, jamal2012poetry, ahmed2019classification, ahmad2020classification}, or from an unsupervised one when there is no prior information about the categories \parencite{navarro2018poetic, ullrich2017relation, barbado2020disco}. 

On many situations, the labelled data available is scarce, not being enough to train a supervised ML model, but being enough to justify using it versus following a fully unsupervised approach. Here, the literature proposes the usage of semi-supervised learning techniques \parencite{van2020survey}, which has been already proven useful for non-poetic text classification \parencite{tian2018semi, giasemidis2018semi, yu2019cvae}. However, there is a lack of analyses that study the usage of semi-supervised learning approaches for poetry text classification, particularly with Spanish poems.

For any of these tasks, poems need to be modelled first through several features. This was traditionally approached by using lexical or affective information \parencite{barros2013automatic, aryani2016measuring}, by word models \parencite{jamal2012poetry, kumar2014poem}, and more recently through the usage of transformers \parencite{ahmad2020classification, de2020predicting}. However, poems are rich texts, and only using one type of features may not be enough to represent their meaning. However, the literature on poetic text modelling also lacks studies regarding this.

Our research focuses on these two aspects: using a labelled corpus of Spanish sonnets and follow a semi-supervised learning approach for inferring the categories of a whole poem corpus, and analyse the importance of combining both lexico-semantic and affective features with State of the Art transformers through sentence embeddings. We use the labelled corpus from \parencite{barbado2020disco} in order to predict the psychological, affective and lexico-semantic categories of a corpus of 4572 sonnets. 

The contributions of the paper are:
\begin{itemize}
    \item Propose a semi-supervised learning framework that infers the psychological, lexico-semantic and affective categories of Spanish sonnets by using affective and lexico-semantic features along with sentence embeddings.
    \item Provide a benchmark for Spanish poetry modelling for both binary and multiclass classification. We use the 21 binary psychological categories and the 10 multiclass affective and lexico-semantic categories from "Diachronic Spanish Sonnet Corpus with Psychological and Affective Labels" (DISCO PAL) corpus \parencite{barbado2020disco}.
    \item Analyse quantitatively the importance of combining lexico-semantic and affective features along with transformers.
    \item Evaluate extending the corpus from the original sonnets used for DISCO PAL (sonnets from XVth to XIXth century), to other modern sonnets (sonnets from XXth century).
\end{itemize}

The remainder of tha paper is organized as follows: Section 2 presents the related literature, focusing on poetry ML classification, as well as the importance of predicting these categories in poems for several domains. Section 3 describes the corpus used, as well as the methodology followed. Section 4 presents the empirical results. Finally, Section 5 shows the conclusions and highlights several research lines that can be pursued.

\section{Related Work}
In this Section, we present related work regarding the modelling of poems for inferring their general meaning, along with the context of our research, particularly to highlight the importance of identifying lexico-semantic, affective and psychological categories within poetry.

\subsection{Poetry Classification}
Poetry express concepts, themes and motifs that can be modelled through its content, similarly to what is done with other literary compositions, through text mining techniques \parencite{liu2012sentiment, miall1994beyond}. A common approach in the literature is through supervised techniques when there is prior knowledge about the poem themes or general meaning. This can be achieved by using Machine Learning (ML) supervised methods.

This approach appears for English poetry in \parencite{lou2015multilabel}, where themes "Love", "Nature", "Social Commentaries", "Religion", "Living", "Relationships", "Activities", "Art \& Sciences", and "Mythology \& Folklore", are predicted through supervised ML in a corpus of 7214 poems. The authors use a Support Vector Machine (SVM) algorithm to model the relationship between Term-Frequency Inverse-Document-Frequency (TD-IDF) features, as well as Latent Dirichlet Allocation (LDA), to represent the poem content and train the ML model. This approach is similar in other languages. In \parencite{jamal2012poetry}, the authors also use a SVM to classify 1500 Malaysian poems into 10 categories. The features used for modelling the poems are also based on TF-IDF. In contrats to this approach, in \parencite{kaur2017punjabi} classify 240 Punjabi poems into 4 categories, benchmarking different ML supervised models (such as AdaBoost), where the input features are lexical variables. In \parencite{ahmed2019classification} authors also benchamrk several ML models (SVM, Naive Bayes, and Linear Support Vector Cassification) for classifying Arabic poetry (92 poems, with a number of verses between 600 and 500) into four categories ("Love", Religious/Islamic", "Politic", "Social"). s input features, they use Boolean vector models, instead of other word counting ones. Besides inferring the category for the whole poem, other approaches focus only on a part of it. In \parencite{kumar2014poem} shows an approach for verse classification in Basque poetry with 212 strophes, considering 6 categories. As input features, they use a Bag of words (BOW) model. In \parencite{de2020predicting}, authors predict the metric pattern category of Spanish poem verses using transformers, together with a supervised learning approach. Focusing on Spanish poetry, \parencite{barros2013automatic} uses a supervised ML model (Decision Tree) for classifying Quevedo's poems (with a corpus of 185 poems) into 4 classes, "Love", "Lisi", "Satiric", and "Philosophical, Moral and Religious (PMR)". As input features for modelling the poems, the authors identified words related to particular emotions (such as "joy", "sadness"...), counting their occurrences within the poems for building the features. More recent researches focus on the usage of Deep Learning (DL). In \parencite{ahmad2020classification} authors use a Long Short-Term Memory (LSTM) based architecture, together with word embeddings, for predicting 13 emotional states in a corpus of 9142 poems. 

The aforementioned studies either have a large labelled corpus, or they work with a small corpus of poems. However, to the best of our knowledge, there are no prior studies of semi-supervised approaches for poetry modelling, specially when we consider Spanish poems. Also, they either focus on transformers for feature generation, or they use lexico-semantic features alone. They do not consider together affective and lexico-semantic features along with transformers.

\subsection{Poetry for Pedagogical and Therapeutic Purposes}
Poetry therapy is the intentional application of written and spoken words for growth and healing, using a language that is condensed, replete with sensory images, and charged with meaning \parencite{gorelick2005poetry}. Its process involves an interaction among three key elements: the poem, the therapist, and the patient/user. The use of literature as a key element in the therapeutic context is a distinguishing characteristic of poetry therapy. \parencite{hynes2019bibliotherapy} state that the literature plays an important role in the therapeutic process by serving as a catalyst. \parencite{hedberg1997re} highlights the cognitive benefits of poetry therapy, demonstrating the ability of poetry to communicate learning, humor, insight and improve verbalizing feeling. Literature addresses how the practice of poetry as a therapy can promote psychological well-being or happiness, based on the pillars of the so-called PERMA model (positive emotion, engagement, relationships, meaning, and accomplishment) \parencite{seligman2011flourish}.

Because of that, poetry therapy has been proven beneficial in several fields, such as psychology, psychiatry, or psychotherapy, as an interdisciplinary treatment method with practical applications, as studied by \parencite{leedy1970value} and \parencite{silverman1983poetry}. \parencite{silverman1988creative} analyses its use in psychotherapy and counselling, \parencite{mcardle2001fiction} explain how literature benefits mental health using expressive writing, bibliotherapy, and poetry therapy. \parencite{lerner1997look} explore the concept “Poetry fixes me”, where poetry provides healing effects. Research also exists in specific areas, as the application of poetry therapy in elderly care and terminally ill patients \parencite{tegner2009evaluating}, where poetry can help enabling the expression of individuals' deepest unspoken concerns within the field of palliative care \parencite{robinson2004personal}, as well as in other sociocultural contexts \parencite{sharma2020writing}. 

Poetry is also beneficial for education, being useful for enhancing learning in of all the development phases. For childhood, it helps developing language skills, while deepening the message learnt, thanks to poetic elements such as the rhyme that create an awareness of the child's own capacity in such a way that it helps to link the words with the relationship of their meaning \parencite{samuelsson2009art}. For old age, poetic language allows to awaken memories related to the affections and feelings experienced by the person, providing the cognitive and affective stimuli of the brain \parencite{aadlandsvik2008second}.

With that, we see that the work carried out in this paper has relevance in several fields; having an extensive corpus of poetry, with identified topics related to psychology, may be useful for applications like poetry therapy.

\section{Method}
In this Section, we present the sonnet corpus used, along with the semi-supervised modelling approach that we propose for inferring the values of several lexico-semantic, affective and psychological concepts evoked by the individual sonnets.

\subsection{Corpus Description}
The corpus used within this paper is a combination from three different input corpora. First, we use the corpus from DISCO (Diachronic Spanish Sonnet Corpus) \parencite{ruiz2018disco}, which includes 4085 sonnets from 15th to 19th centuries. This is combined with the corpus DISCO PAL (Diachronic Spanish Sonnet Corpus with Psychological and Affective Labels) from \parencite{barbado2020disco}, a subset of 274 sonnets from DISCO with expert annotations for the following affective, lexico-semantic and psychological categories:

\begin{table}[h!]
\centering
\resizebox{140pt}{!}{%
\begin{tabular}{@{}lll@{}} 
\toprule
& Affective categories & \\ 
\midrule
 Valence & & Arousal \\
 Happiness & & Disgust\\ 
 Anger & & Sadness \\
 Fear & & \\ [1ex] 
\bottomrule
\end{tabular}%
}
\caption{Affective categories used with a scale of 1 to 4, using integer values.}
\label{table:affective-list}
\end{table}

\begin{table}[h!]
\centering
\resizebox{200pt}{!}{%
\begin{tabular}{@{}lll@{}} 
\toprule
& Lexico-semantic categories & \\ 
\midrule
 Concreteness & & Imageability\\
 Context availability & & \\ [1ex] 
\bottomrule
\end{tabular}%
}
\caption{Lexico-semantic categories used with a scale of 1 to 4, using integer values.}
\label{table:lexico-semantic-list}
\end{table}

\begin{table}[h!]
\centering
\resizebox{230pt}{!}{%
\begin{tabular}{@{}lll@{}} 
\toprule
& Psychological categories & \\ 
\midrule
Solitude (Soledad) & & Anxiety (Ansiedad)\\ 
Illusion (Ilusión) & & Anger/Wrath (Ira)\\
Daydream (Ensoñación) & & Instability (Inestabilidad)\\
Grandeur (Grandiosidad) & & Idealization (Idealización)\\
Pride (Orgullo) & &  Depression (Depresión)\\ 
Irritability (Irritabilidad) & & Disappointment (Desilusión)\\
Dramatisation (Dramatización) & & Prejudice (Prejuicio)\\
Aversion/Loathing (Aversión) & & Insecurity (Inseguridad)\\
Helplessness (Impotencia) & & Vulnerability (Vulnerabilidad)\\ 
Fear (Temor) & & Obsession (Obsesión)\\
Compulsion (Compulsión) & & \\[1ex] 
\bottomrule
\end{tabular}%
}
\caption{Psychological categories used with a scale of 0 to 1 (binary values).}
\label{table:psycho-list}
\end{table}

\begin{figure}[h!]
\centering
\begin{tabular}{c@{\qquad}c@{\qquad}c}
\includegraphics[width=\textwidth]{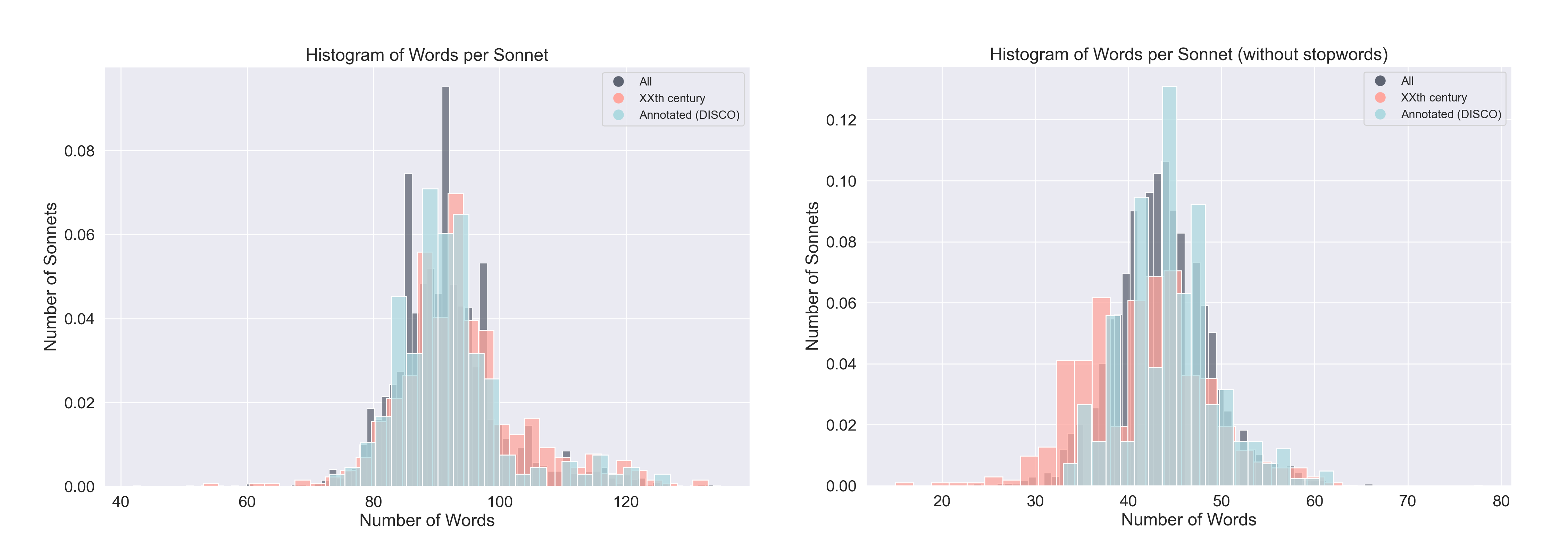}
\end{tabular} 
\caption{Histogram of words, with and without stopword, for each sonnet considering the whole corpus, the subset of new sonnets from S.XX, and the annotated subset from \parencite{barbado2020disco}.}
\label{fig:histogram_words_corpora}
\end{figure}

Finally, we expand the corpus by including 532 sonnets from \parencite{amediavoz2013} with authors from the XXth century with Spanish as mother tongue.  
The corpora DISCO and DISCO PAL contains some multiple-part sonnets. For this paper, we will only focus on single-part sonnets (2 quartets and 2 tercets, for a total of 14 lines in 4 stanzas). This means using 270 sonnets from DISCO PAL and 4040 from DISCO. Thus, the total number of sonnets used are 4572, with 270 annotated with expert domain knowledge. Figure \ref{fig:histogram_words_corpora} shows the histogram of words for our new corpus, with and without stopwords (M=91.8, Std=8.7; M=43.5, Std=5.2), compared to the subset of XXth century (M=94, Std=10.3; M=41.6, Std=6.7) and to the subset annotated from DISCO PAL (M=92.2, Std=8.73; M=44.7, Std=5.0).

\begin{figure}[h!]
\centering
\begin{tabular}{c@{\qquad}c@{\qquad}c}
\includegraphics[width=250pt]{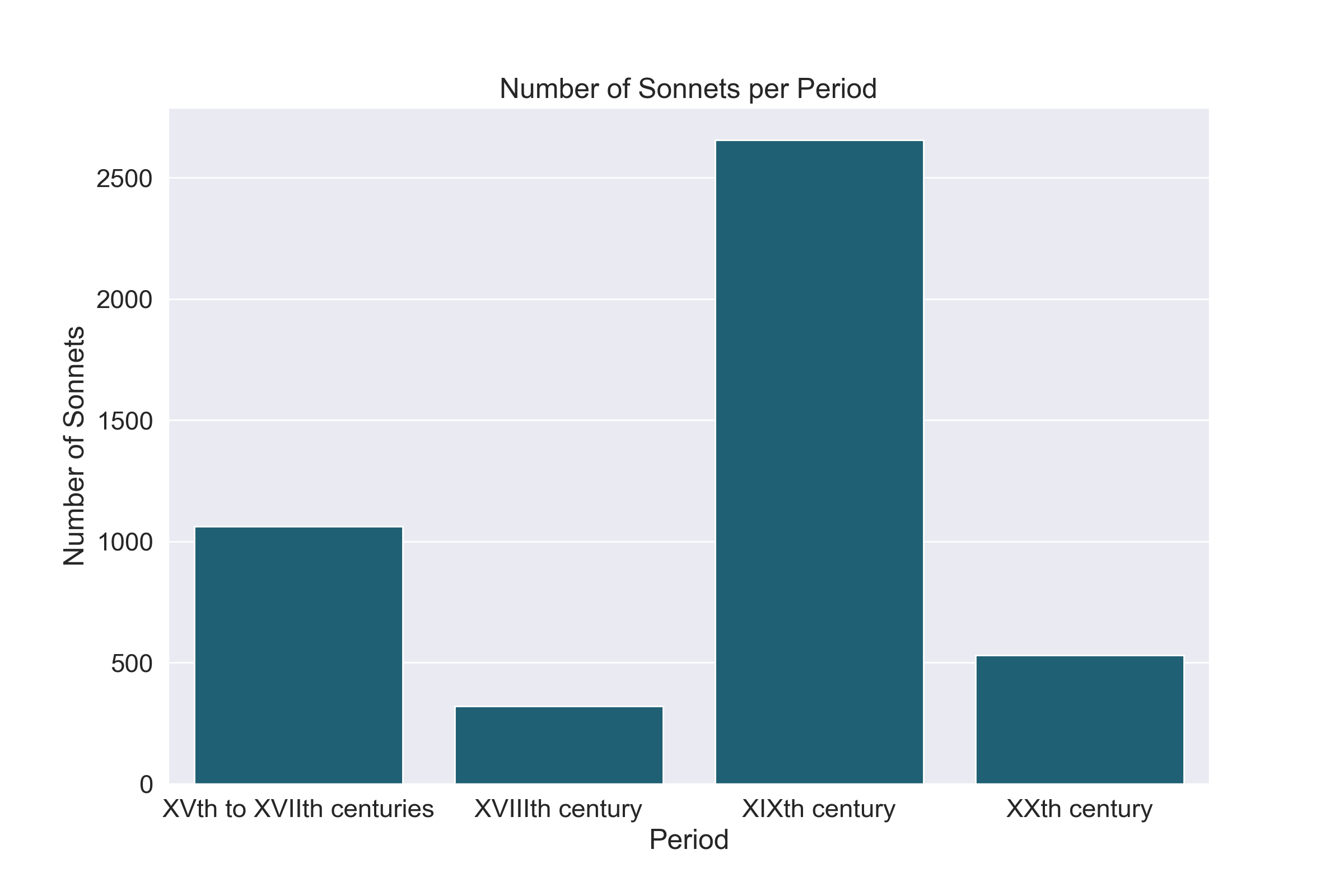}
\end{tabular} 
\caption{Sonnets per time period, including the new ones for S.XX.}
\label{fig:sonnets_per_period}
\end{figure}

Both the number of sonnets per psychological category (with "Prejudice" and "Obsession" the categories with fewer sonnets, 30 and 32 respectively), as well as the number of sonnets from XXth included, are enough from the a statistical point of view based on a power analysis with an alpha of 0.1, a Cohen’s d of 0.8 and the default statistical power of 0.8 (which sets the minimum in 20) \parencite{cohen1992power, sullivan2012using}. The number of sonnets is also higher when compared to other previous works related to poetry, such as \parencite{ullrich2017relation, aryani2016measuring}, where the authors work with the categories “friendliness”, “sadness” and “spitefulness” and they are associated to 19, 21 and 17 poems respectively. Figure \ref{fig:sonnets_per_period} shows the number of sonnets per period, including the new sonnets from XXth century.

\subsection{Semi-supervised Modelling}
\begin{figure}[h!]
\centering
\begin{tabular}{c@{\qquad}c@{\qquad}c}
\includegraphics[width=320pt]{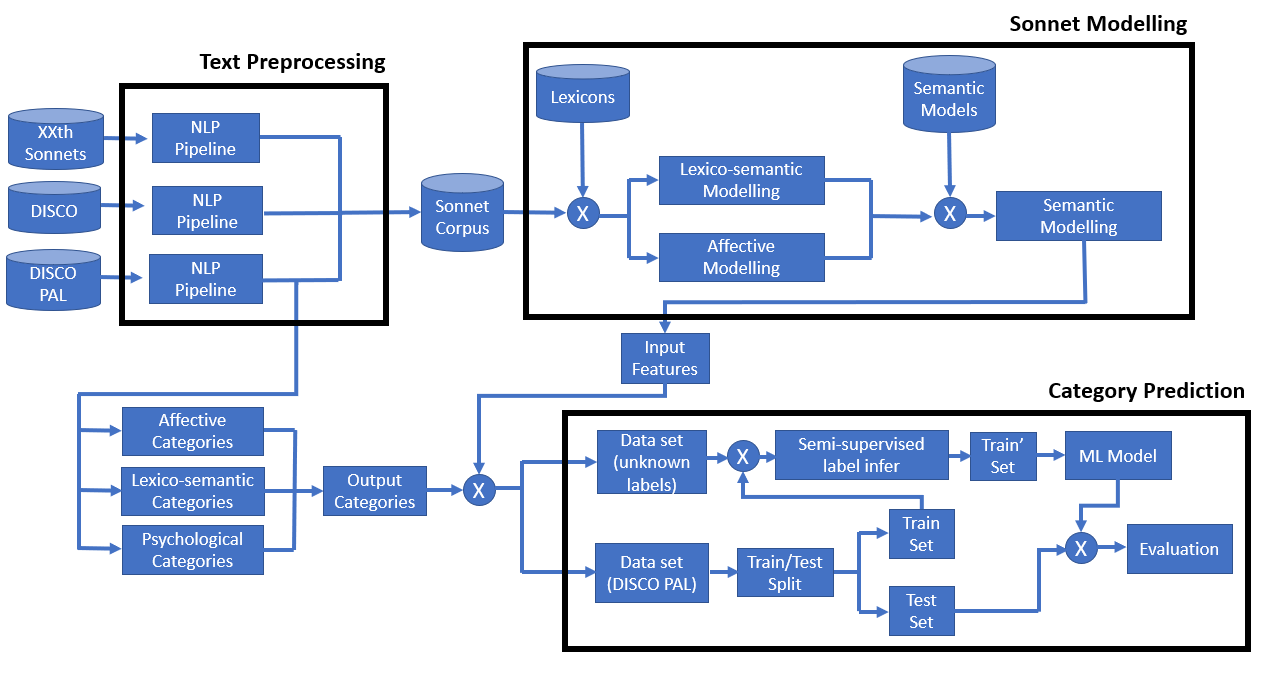}
\end{tabular} 
\caption{Flowchart followed by the process, with three main phases: an initial text processing, the lexico-semantic and affective modelling of the sonnets, and the category prediction using semi-supervised learning.}
\label{fig:flowchart-process}
\end{figure}

With the corpus described in Subsection 3.1, we follow the flowchart described in \ref{fig:flowchart-process} in order to infer the categories from Tables \ref{table:affective-list} and \ref{table:lexico-semantic-list} for the whole corpus using the information available at DISCO PAL.

The first part of the flowchart is a NLP (natural language processing) pipeline that performs the following steps: lemmatization and part-of-speech tagging, tokenization, elimination of stopwords and stemming. 

The second phase of the flowchart includes all the steps for the lexico-semantic and affective modelling of the sonnets from the corpus, in order to infer its General Affective Meaning (GAM) \parencite{}. These features are the ones from \parencite{barbado2020disco} since they have been already proved as useful for modelling the GAM of Spanish sonnets. They are similar to the ones used in other works, such as \parencite{ullrich2017relation} for inferring the GAM on German poetry. The features are described in Annex (A.2).
These features are obtained using several Spanish lexicons that provide lexico-semantic and affective values for individual words. In particular, we use \parencite{perez2021emopro},  \parencite{stadthagen2018norms}, \parencite{hinojosa2016affective}, \parencite{guasch2016spanish} and \parencite{ferre2017moved}. These lexicons are combined into one, where we keep the mean value for each word in case there are duplicates. The lexicons do not need rescaling since a features have the same ranges (valence, arousal, concreteness, imageability and context availability features have a range of 1 to 9; happiness, anger, sadness, fear and disgust features have a range of 1 to 5).

\begin{figure}[h!]
\centering
\begin{tabular}{c@{\qquad}c@{\qquad}c}
\includegraphics[width=320pt]{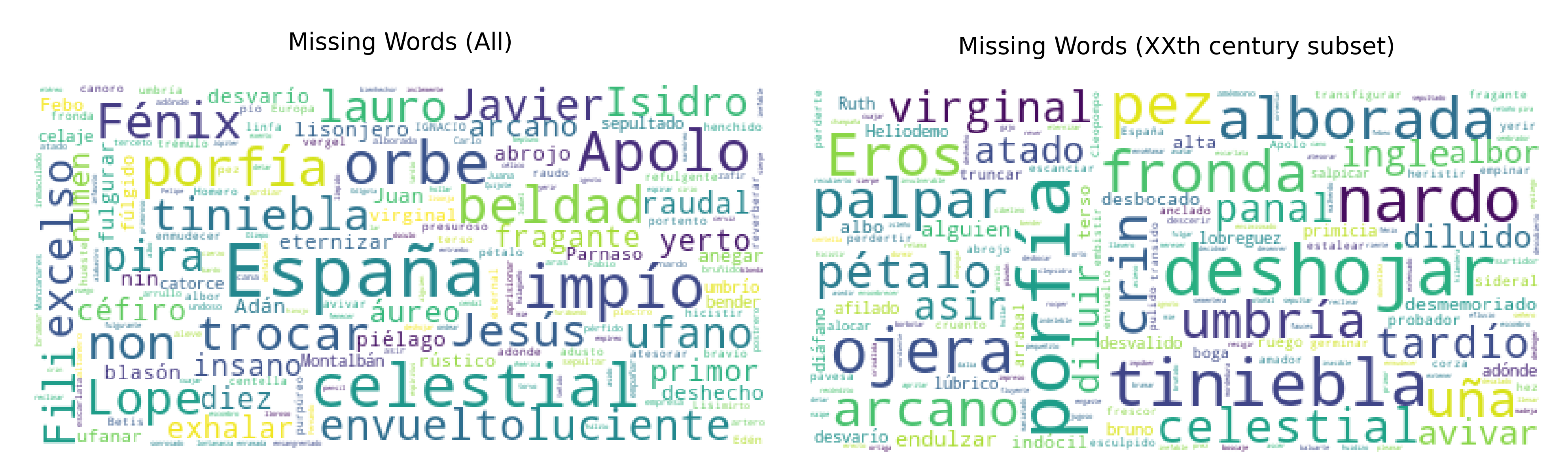}
\end{tabular} 
\caption{Missing from all lexicons, considering either the full corpus or the subset of sonnets from XXth century.}
\label{fig:wordcloud_missing}
\end{figure}

In order to assign lexico-semantic and affective values to the individual words for building the features from (A.2), we use the stemmed words in the sonents (after the text preprocessing phase) and we match them with the corresponding stemmed lexicon entry. If there are several possible entries, we assign the mean value of all of them. We use the stemmed version of the words for crossing the two data source since it provides the best coverage of sonnet words, as Table \ref{table:percentage-words-per-lexicon} shows. The total word coverage is 47\%, with some words missing within the lexicon, as can be seen in Figure \ref{fig:wordcloud_missing}, representing the word cloud of missing the sonnets of XXth century, as well as on the full corpus. Most of the missing words are because some sonnets are written in old Spanish. This is the reason why the percentage of coverage increases when we consider only XXth. The percentage for stemmed words within this subset is 74\%, and it increases to 93\% if we consider the relative appearances of the words, highlighting that in many cases the missing words are hapaxes.

\begin{table}[h!]
\centering
\resizebox{\textwidth}{!}{%
\begin{tabular}{@{}llllll@{}}
\toprule
\textbf{Lexicon} &
  \textbf{Processing} &
  \textbf{\begin{tabular}[c]{@{}l@{}}Corpus\\ (All)\end{tabular}} &
  \textbf{\begin{tabular}[c]{@{}l@{}}Corpus\\ (DISCO)\end{tabular}} &
  \textbf{\begin{tabular}[c]{@{}l@{}}Corpus\\ (XXth)\end{tabular}} &
  \textbf{\begin{tabular}[c]{@{}l@{}}Corpus\\ (Annotated)\end{tabular}} \\ \midrule
All Lexicons                                       & Stemmed    & 0.47 & 0.48 & 0.74 & 0.73 \\
All Lexicons                                       & Lemmatized & 0.39 & 0.4  & 0.63 & 0.64 \\
All Lexicons                                       & Original   & 0.2  & 0.21 & 0.34 & 0.36 \\
\parencite{perez2021emopro}       & Stemmed    & 0.05 & 0.05 & 0.10 & 0.11 \\
\parencite{perez2021emopro}       & Lemmatized & 0.04 & 0.04 & 0.08 & 0.09 \\
\parencite{perez2021emopro}        & Original   & 0.02 & 0.02 & 0.04 & 0.05 \\
\parencite{stadthagen2018norms}   & Stemmed    & 0.37 & 0.37 & 0.57 & 0.55 \\
\parencite{stadthagen2018norms}    & Lemmatized & 0.26 & 0.26 & 0.38 & 0.37 \\
\parencite{stadthagen2018norms}    & Original   & 0.14 & 0.14 & 0.21 & 0.21 \\
\parencite{hinojosa2016affective} & Stemmed    & 0.05 & 0.05 & 0.11 & 0.12 \\
\parencite{hinojosa2016affective} & Lemmatized & 0.03 & 0.04 & 0.08 & 0.08 \\
\parencite{hinojosa2016affective} & Original   & 0.02 & 0.02 & 0.03 & 0.04 \\
\parencite{guasch2016spanish}     & Stemmed    & 0.08 & 0.08 & 0.16 & 0.15 \\
\parencite{guasch2016spanish}     & Lemmatized & 0.05 & 0.06 & 0.10 & 0.1  \\
\parencite{guasch2016spanish}     & Original   & 0.03 & 0.03 & 0.06 & 0.07 \\
\parencite{ferre2017moved}        & Stemmed    & 0.12 & 0.13 & 0.23 & 0.25 \\
\parencite{ferre2017moved}         & Lemmatized & 0.09 & 0.09 & 0.16 & 0.18 \\
\parencite{ferre2017moved}        & Original   & 0.05 & 0.05 & 0.10 & 0.11 \\ \bottomrule
\end{tabular}%
}
\caption{Percentage of words from the corpus (without stopwords) within each of the lexicons, considering original words, words after lemmatization and words after stemming.}
\label{table:percentage-words-per-lexicon}
\end{table}

After the lexico-semantic and affective modelling steps, the sonnet modelling phase includes a semantic modelling step based on word embedding techniques. Particularly, since sonnets are highly-structured and short text compositions, our proposed approach is modelling them through pre-trained sentence transformers. This can be done since pre-trained models for sentence embeddings are normally fine-tuned for small text compositions (up to 128 words \parencite{reimers-2020-multilingual-sentence-bert}), and within our corpus, sonnets do not exceed 65 word pieces (without stopwords). The pre-trained models that we evaluate for modelling the corpus of sonnets are the multi-lingual models shown below \parencite{reimers-2019-sentence-bert}:

\begin{itemize}
    \item \textit{quora-distilbert-multilingual}: Model using DistilBertModel as transformer, mean pooling, with 768 features.
    \item \textit{stsb-xlm-r-multilingual}: Model using XLMRobertaModel as transformer, mean pooling, with 768 features.
    \item \textit{paraphrase-multilingual-mpnet-base-v2}: Model using XLMRobertaModel  as transformer, mean pooling, with 768 features.
    \item \textit{paraphrase-xlm-r-multilingual-v1}: Model using Distil-Roberta-Base  as transformer, mean pooling, with 768 features.
    \item \textit{distiluse-base-multilingual-cased-v1}: Model using the Universal Sentence Encoder,  mean pooling, with 512 features. 
\end{itemize}

These sentence embedding models are complemented with our proposed sentence embedding that weights the words according to their affective meaning. Particularly, we normalize the lexicon and associate as a weight for each word the maximum normalized value among all the lexico-semantic and affective features. Then, we apply a mean weighted pooling according to those weights. This is shown in Equation \ref{eq:aff-weighted-pooling}, with $N$ the number of words, $\vec{v}(k)$ the word embedding for word $k$, $w(k)$ the word weight. This $w(k)$ value corresponds to the maximum normalized feature value from (A.2).

\begin{equation}\label{eq:aff-weighted-pooling}
\begin{aligned}
\vec{v}_n = \frac {\sum_{k=1}^{N} \vec{v}(k) \times w(k)} {N}
\\
\text{with   } w(k) = max(f_1(k), f_2(k),...,k_m(k))
\end{aligned}
\end{equation}

For the individual word embeddings, we use:

\begin{itemize}
    \item \textit{bert-base-multilingual-cased}: Pretrained multilingual based model, trained over Wikipedia texts, with 768 features  \parencite{DBLP:journals/corr/abs-1810-04805}.
    \item \textit{dccuchile/bert-base-spanish-wwm-cased}: BETO, Spanish version of BERT, which has been seen useful in some domains when compared to multilingual models \parencite{CaneteCFP2020}. 768 features.
\end{itemize}

The final phase consists in inferring the lexico-semantic, affective and psychological categories through the modelling of sonnets through lexico-semantic and affective features. \parencite{barbado2020disco} already proposes using the mean values from the input features as GAM for the sonnet. However, we will see if the inferring process through semi-supervised learning improves it.
In our approach, we select a random small subset of sonnets from DISCO PAL (in particular, we sample a maximum of two sonnets per category for building the train set), using them to for the semi-supervised learning task, together with the rest of the corpus. For tackling the semi-supervised learning task, we consider the following approaches:
\begin{itemize}
    \item Combining the algorithm from \parencite{yarowsky1995unsupervised} together with LightGBM (Light Gradient Boosting Machine) algorithm \parencite{ke2017lightgbm}. The reason behind it is that this semi-supervised learning algorithm is a self-training model that can be combined with any supervised Machine Learning algorithm, and LightGBM is an algorithm well-known for its good performance over imbalanced data sets, such as our sonnet corpus (where the annotated categories are imbalanced).
    \item Using the semi-supervised LabelSpreading algorithm \parencite{zhou2004learning}, which infers the missing labels by finding nearby points in the space and assigning them the same label as the data points with known labels. This can be done by either using a K-Nearest Neighbors (KNN) approach, or a Radial Basis Function (RBF) one.
    \item Using the semi-supervised learning output from LabelSpreading as a pre-training phase for training a supervised algorithm \parencite{van2020survey}. In our case, the supervised ML algorithm is LightGBM. This offers also the possibility of using over-sampling techniques over the whole training data. In our case, we will analyse the results by using SMOTE (Synthetic Minority Over-Sampling Technique) \parencite{chawla2002smote}
\end{itemize}

With that, an individual semi-supervised model will be trained for predicting each of the categories mentioned in Subsection 3.1.

\section{Results}
In this Section, we present the results of the different experiments that we have carried out in order to answer the following research questions:
\begin{itemize}
    \item \textbf{Q1}: \textit{Is the proposal able to accurately infer the lexico-semantic, affective and psychological categories over the whole corpus?}
    \item \textbf{Q2}: \textit{Are the lexico-semantic and affective features significantly important for predicting the different categories?}
    \item \textbf{Q3}: \textit{Are the results significantly different if we use only the original DISCO sonnets (since the annotated sonnets are a subcorpus from DISCO) compared to the results obtained by using additional sonnets (XXth sonnets)?}
\end{itemize}

For \textbf{Q1}, we focus on three well-known classification metrics in order to assess the quality of the predictions.

First, we consider F1 metric. This metric is common within classification tasks, and useful when the labels are imbalanced \parencite{jeni2013facing}. \parencite{lou2015multilabel} uses binary classifiers to identify on of the 9 possible category associated to the poems. Using this as a reference benchmark, the results on F1 range from \textbf{0.744} to \textbf{0.948} on their best run (with an average of  \textbf{0.835}). As a reference for Spanish language models (but for prose texts), \parencite{alhuzali2021spanemo} shows SOTA benchmark metrics for multiclass emotion text classification. For Spanish language models, F1 is between \textbf{0.532} (F1-macro) and \textbf{0.641} (F1-micro). 

We also use the Cohen's Kappa metric, that measures the agreement between one annotator (the ground truth and another annotator (the predictions) \parencite{cohen1960coefficient}. This metric is useful for both comparing models (and choosing the one with better kappa), as well as for seeing if the results are good by themselves, since it offers thresholds to evaluate the results. \parencite{landis1977application} propose the following ones: $kappa < 0$: Poor agreement, $0 < kappa < 0.20$: Slight agreement, $0.21 < kappa < 0.40$: Medium (moderate) agreement, $0.41 < kappa < 0.60$: Sufficient (fair) agreement, $0.61 < kappa < 0.80$: Considerable agreement, and $0.81 < kappa < 1.0$: Almost perfect agreement.

Third, we use the Area Under the Curve (AUC) as a metric to assess the quality of the explanations. Though this metric depends on the context and not every domain considers the same thresholds to identify when the values are good, on general terms the following reference values can be considered \parencite{hosmer2013applied}: $AUC = 0.5$: No discrimination, $0.5 < AUC < 0.7$: Poor discrimination, $0.7 \geq AUC < 0.8$: Acceptable discrimination, $0.8 \geq AUC < 0.9$: Excellent discrimination and $0.9 \geq AUC$: Outstanding discrimination. Using again the reference of \parencite{lou2015multilabel} for poetry in English, the AUC ranges on their best run from \textbf{0.628} to \textbf{0.766} (with an average of \textbf{0.700}).


For \textbf{Q1}, \textbf{Q2} and \textbf{Q3} we apply a Random Cross Validation approach (CV) in order to check the results over different runs (Q1), when using affective features versus not using them (Q2), and when using only DISCO versus using also external sonnets (Q3). We use 20 random samples, since the value is enough from the a statistical point of view based on a power analysis with an alpha of 0.1, a Cohen’s d of 0.8 and the default statistical power of 0.8 (which sets the minimum in 20).


\subsection{Nomenclature}
Here we indicate the nomenclature we use for the different combination of algorithms that we use within our analyses.
\begin{itemize}
    \item LS-LightGBM: Combination of LabelSpreading with LightGBM. It can either be LS-LightGBM-KNN (KNN kernel) or LS-LightGBM-RBF (RBF kernel). If it includes SMOTE, then we refer to it as LS-LightGBM-SMOTE-KNN or LS-LightGBM-SMOTE-RBF.
    \item ST-LightGBM: Self-Training algorithm with LightGBM.
\end{itemize}

\subsection{Data Description}
For the analyses, we consider a Random Cross Validation (CV), where we use a subsample of sonnets for training an individual ML model on for predicting each category, and then we evaluate the results over the remaining sonnets. We choose the random subset of sonnets by selecting two sonnets from each category value This leaves an average of 28\% sonnets for training, and the remaining 72\% for evaluation.

\subsection{Psychological Categories}
Here, we analyse Q1, Q2 and Q3 regarding the binary psychological categories. We first check Q1 by choosing the semantic transformers and semi-supervised models (introduced in Subsection 3.2) that yield the best metric results over the Random CV. Table \ref{table:semisupervised-psycho-best-cv} shows the best semantic and predictive models for each category, along with their mean metric values. It also shows the mean data distribution of the test set for each category, where we see that in all cases, except for 'Prejudice', the number of sonnets with that category are above the reference of 20. Regarding F1, we see that 9 categories have values above the reference 0.744 considered. For Cohen's Kappa, 9 categories have values belonging to the "moderate agreement" group; the rest of the categories have a "slight agreement". Finally, for AUC, 18 categories have values above the minimum reference of 0.628. Among those categories, 8 of them have values within the "acceptable discrimination" general reference. In particular, the following categories have at least two metrics with values above the SOTA references: "Aversion", "Depression", "Disappointment", "Dramatisation", "Daydream", "Grandeur", "Idealization", "Illusion", "Instability", "Insecurity", "Anger (binary)", "Irritability", "Pride", "Prejudice", "Solitude" and "Vulnerability" (76\% of the categories). 

\begin{table}[h!]
\centering
\resizebox{\textwidth}{!}{%
\begin{tabular}{@{}llllllll@{}}
\toprule
\textbf{Category} &
  \textbf{Semantic Model} &
  \textbf{Predictive Model} &
  \textbf{\begin{tabular}[c]{@{}l@{}}N Class 0\\ Test\end{tabular}} &
  \textbf{\begin{tabular}[c]{@{}l@{}}N Class 1\\ Test\end{tabular}} &
  \textbf{F1 (weighted)} &
  \textbf{Kappa} &
  \textbf{AUC} \\ \midrule
Anxiety        & paraphrase-multilingual-mpnet-base-v2                      & ST-LightGBM               & 146.0  & 51.25 & 0.69 & 0.17 & 0.66 \\
Aversion       & stsb-xlm-r-multilingual                      & ST-LightGBM               & 130.4 & 66.9  & 0.69 & 0.32 & 0.74 \\
Compulsion     & bert-base-multilingual-cased     & ST-LightGBM               & 159.3 & 38.0  & 0.72 & 0.06 & 0.55 \\
Depression     & paraphrase-multilingual-mpnet-base-v2                      & ST-LightGBM               & 172.5  & 24.8 & 0.83 & 0.18 & 0.71 \\
Disappointment & paraphrase-multilingual-mpnet-base-v2                      & ST-LightGBM               & 167.3 & 30.0  & 0.78 & 0.13 & 0.64 \\
Dramatisation  & paraphrase-multilingual-mpnet-base-v2                      & ST-LightGBM               & 122.8 & 74.5  & 0.63 & 0.22 & 0.66 \\
Daydream       & quora-distilbert-multilingual                      &  LS-LightGBM-SMOTE-KNN & 164.6 & 32.7  & 0.78 & 0.15 & 0.7  \\
Grandeur       & stsb-xlm-r-multilingual                      & ST-LightGBM               & 118.5 & 78.8  & 0.67 & 0.31 & 0.74 \\
Idealization   & bert-base-spanish-wwm-cased &  LS-LightGBM-SMOTE-KNN & 119.3 & 78.0  & 0.66 & 0.3  & 0.7  \\
Illusion       & paraphrase-multilingual-mpnet-base-v2                      &  LS-LightGBM-SMOTE-KNN & 144.3  & 52.9 & 0.71 & 0.26 & 0.72 \\
Helplessness    & quora-distilbert-multilingual                      & LS-LightGBM-SMOTE-RBF & 156.6 & 40.7  & 0.69 & 0.19 & 0.67 \\
Instability   & bert-base-multilingual-cased     &  LS-LightGBM-SMOTE-KNN & 154.6 & 42.7  & 0.72 & 0.21 & 0.68 \\
Insecurity     & distiluse-base-multilingual-cased-v1                      & ST-LightGBM               & 167.7 & 29.6  & 0.79 & 0.08 & 0.65 \\
Anger (binary)        & stsb-xlm-r-multilingual                      & ST-LightGBM               & 160.8 & 36.5  & 0.75 & 0.17 & 0.68 \\
Irritability   & quora-distilbert-multilingual                      & ST-LightGBM               & 175.6 & 21.7  & 0.84 & 0.1  & 0.69 \\
Obsession      & quora-distilbert-multilingual                      & LS-LightGBM-SMOTE-RBF & 176.2  & 21.1 & 0.82 & 0.08 & 0.61 \\
Pride          & stsb-xlm-r-multilingual                      & ST-LightGBM               & 144.0  & 53.3 & 0.7  & 0.23 & 0.69 \\
Prejudice      & quora-distilbert-multilingual                      & ST-LightGBM               & 178.9  & 18.4 & 0.87 & 0.1  & 0.74 \\
Solitude       & paraphrase-multilingual-mpnet-base-v2                      & ST-LightGBM               & 154.8 & 42.5  & 0.75 & 0.24 & 0.72 \\
Fear (binary)  & bert-base-multilingual-cased     &  LS-LightGBM-SMOTE-KNN & 131.5  & 65.8 & 0.6  & 0.15 & 0.62 \\
Vulnerability  & bert-base-multilingual-cased     &  LS-LightGBM-SMOTE-KNN & 107.8  & 89.5 & 0.59 & 0.21 & 0.67 \\ \bottomrule
\end{tabular}%
}
\caption{Best combinations of semantic models and semi-supervised techniques for the different psychological categories.}
\label{table:semisupervised-psycho-best-cv}
\end{table}

\begin{figure}[h!]
\centering
  \begin{tabular}{c@{\qquad}c@{\qquad}c}
  \includegraphics[width=300pt]{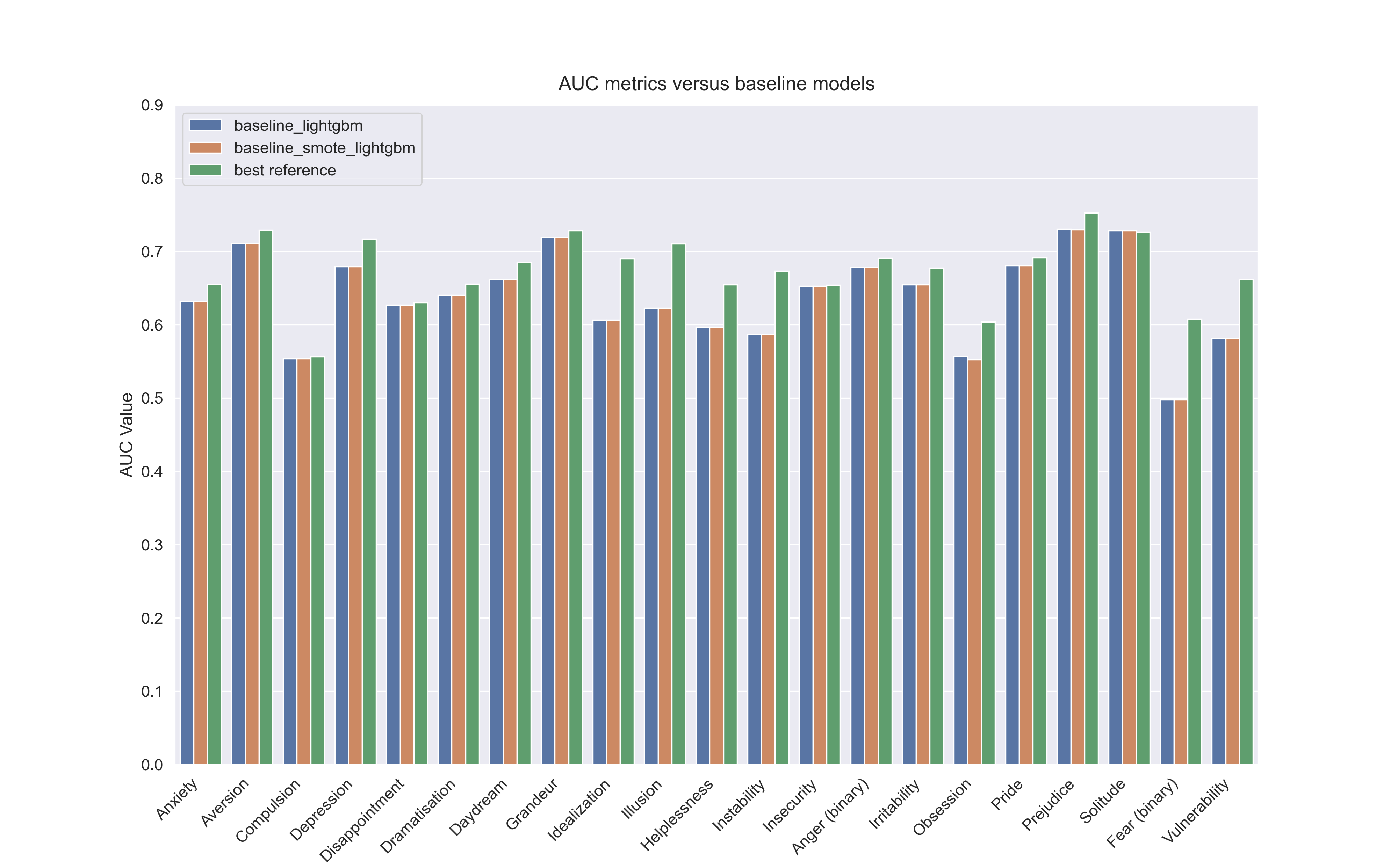}
  \end{tabular} 
  \caption{Comparison of the semi-supervised approach versus baseline models.\label{fig:df_plot_psycho_metrics_vs_baseline}}
\end{figure}

Figure \ref{fig:df_plot_psycho_metrics_vs_baseline} shows the comparison of the AUC value with respect to a baseline LigthGBM model (with and without SMOTE) without using semi-supervised learning. Considering these baseline models, in all the categories, the semi-supervised learning approach is either similar or better than the baselines counterparts. For "Dramatisation", "Obsession", "Depression", "Helplessness", "Anxiety", "Instability", "Idealization", "Fear (binary)", "Vulnerability", and "Illusion", the results are statistically different according to Wilcoxcon signed-rank test \parencite{conover1998practical}, with the semi-supervised approach being better. For the remaining categories, even though the semi-supervised approach improves the baseline models in some cases, the results are not statistically different: "Compulsion" (p-value 0.97), "Disappointment" (p-value=0.96), "Solitude" (p-value=0.89), "Insecurity" (p-value=0.78), "Grandeur" (p-value=0.43), "Pride" (p-value=0.33), "Irritability" (p-value=0.3), "Anger (binary)" (p-value=0.29), "Prejudice" (p-value=0.16), "Daydream" (p-value=0.12), and "Aversion" (p-value=0.11). The results of the semi-supervised models are also, for all the categories, significantly better than those of a random classifier (AUC=0.5).

\begin{figure}[h!]
\centering
  \begin{tabular}{c@{\qquad}c@{\qquad}c}
  \includegraphics[width=300pt]{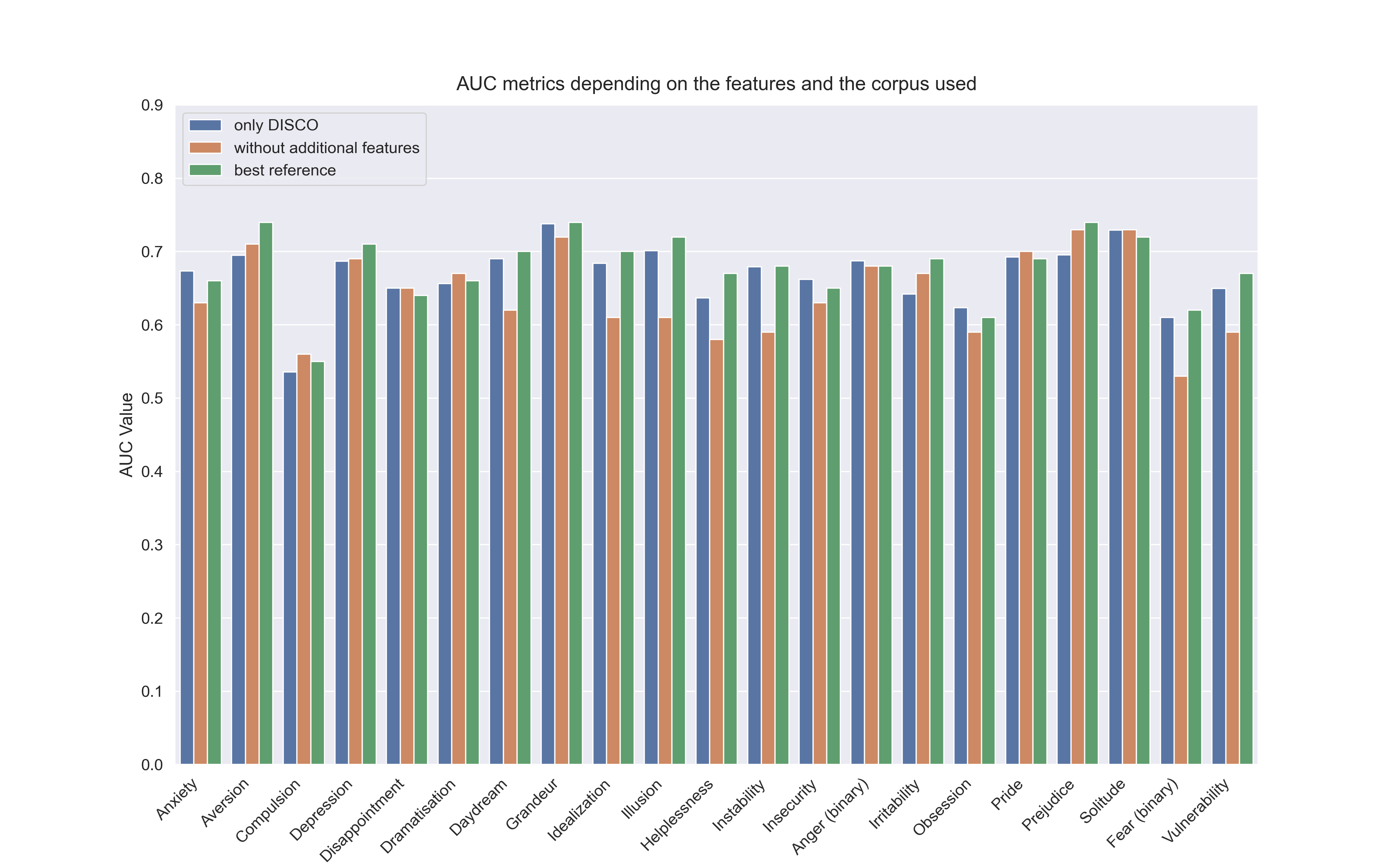}
  \end{tabular} 
  \caption{Comparison of the results without using affective and lexico-semantic additional features, and also when using only DISCO.\label{fig:df_plot_psycho_metrics_q2_q3}}
\end{figure}

Regarding Q2 and Q3, the results are shown in Figure \ref{fig:df_plot_psycho_metrics_q2_q3}. For Q2, there are 11 categories with statistically significant differences according to Wilcoxcon signed-rank test. They are "Anxiety", "Aversion", "Daydream", "Grandeur", "Idealization", "Illusion", "Helplessness", "Instability", "Irritability", "Fear (binary)", and "Vulnerability". In all those cases, the AUC metrics improve when using the affective and lexico-semantic additional features together with the semantic transformers. The AUC improvement in those cases span from 0.02 to 0.12 (with a mean improvement of 0.07).
For Q3, there are only 5 categories where the AUC metrics are statistically significant: "Aversion", "Depression", "Helplessness", "Irritability" and "Prejudice". However, using additional sonnets from the XXth is actually improving the results in all those cases. The AUC improvement in those cases span from 0.02 to 0.05 (with a mean improvement of 0.03).

\subsection{Lexico-semantic and Affective Features}
Similar to the previous section, here we analyse Q1, Q2 and Q3 considering the lexico-semantic and affective categories. In this case, the models are multiclass, since the categories have 4 possible values. Table \ref{table:semisupervised-emotions-best-cv} shows the best semantic and predictive models for each category, along with their mean metric values. It also shows the mean data distribution of the test set for each category.
Regarding F1, we see that only 3 out of the 10 categories have a F1 value above the reference of 0.532 ("anger (ordinal)", "fear (ordinal)", "happiness"). For Cohen's Kappa, only "fear (ordinal)" has a value within the "moderate agreement" group (the remaining categories are within the "slight agreement" one). Finally, for AUC, 6 out of the 10 categories have values above the 0.628 threshold considered (we also use as a reference the AUC used from binary classification models). In particular, the following categories have at least two metrics with values above the SOTA reference: "anger (ordinal)", "fear (ordinal)", and "happiness". "disgust", "sadness" and "valence" have metrics that only have one metric above the SOTA reference (for AUC). "arousal", "concreteness", "context availability", and "imageability" are the categories with worst results, being below the reference values for all the metrics.

\begin{table}[h!]
\centering
\resizebox{\textwidth}{!}{%
\begin{tabular}{@{}llllllllll@{}}
\toprule
\textbf{Category} &
  \textbf{\begin{tabular}[c]{@{}l@{}}N Class 1\\ Test\end{tabular}} &
  \textbf{\begin{tabular}[c]{@{}l@{}}N Class 2\\ Test\end{tabular}} &
  \textbf{\begin{tabular}[c]{@{}l@{}}N Class 3\\ Test\end{tabular}} &
  \textbf{\begin{tabular}[c]{@{}l@{}}N Class 4\\ Test\end{tabular}} &
  \textbf{Semantic Model} &
  \textbf{Predictive Model} &
  \textbf{F1 (weighted)} &
  \textbf{Kappa} &
  \textbf{AUC} \\ \midrule
anger (ordinal)             & 144.2 & 32.9 & 18.5 & 5.3  & stsb-xlm-r-multilingual &  LS-LightGBM-SMOTE-KNN & 0.63 & 0.14 & 0.65 \\
arousal              & 32.0  & 98.6 & 60.3 & 10.0 & paraphrase-multilingual-mpnet-base-v2 &
ST-LightGBM               & 0.43 & 0.09 & 0.59 \\
concreteness         & 78.4  & 54.2 & 45.0 & 23.2 & quora-distilbert-multilingual & 
ST-LightGBM               & 0.32 & 0.06 & 0.55 \\
context availability & 89.0  & 57.6 & 36.4 & 17.9 & paraphrase-xlm-r-multilingual-v1 &  LS-LightGBM-RBF & 0.36 & 0.07 & 0.55 \\
disgust              & 108.6 & 55.9 & 32.4 & 4.0  & stsb-xlm-r-multilingual & 
ST-LightGBM               & 0.5  & 0.16 & 0.66 \\
fear (ordinal)       & 143.9 & 46.8 & 9.4  & 1.0  & paraphrase-multilingual-mpnet-base-v2 & 
ST-LightGBM               & 0.68 & 0.22 & 0.7  \\
happiness            & 134.7 & 41.0 & 20.1 & 5.1  & stsb-xlm-r-multilingual &  LS-LightGBM-KNN & 0.58 & 0.16 & 0.67 \\
imageability         & 91.9  & 54.5 & 39.7 & 14.9 & paraphrase-xlm-r-multilingual-v1 & LS-LightGBM-RBF        & 0.38 & 0.07 & 0.56 \\
sadness              & 67.4  & 56.1 & 55.7 & 21.7 & paraphrase-multilingual-mpnet-base-v2 & 
ST-LightGBM               & 0.4  & 0.2  & 0.67 \\
valence              & 9.8   & 61.5 & 69.8 & 59.8 & stsb-xlm-r-multilingual & 
ST-LightGBM               & 0.42 & 0.19 & 0.65 \\ \bottomrule
\end{tabular}%
}
\caption{Best combinations of semantic models and semi-supervised techniques for the different lexico-semantic and affective categories.}
\label{table:semisupervised-emotions-best-cv}
\end{table}

\begin{figure}[h!]
\centering
  \begin{tabular}{c@{\qquad}c@{\qquad}c}
  \includegraphics[width=300pt]{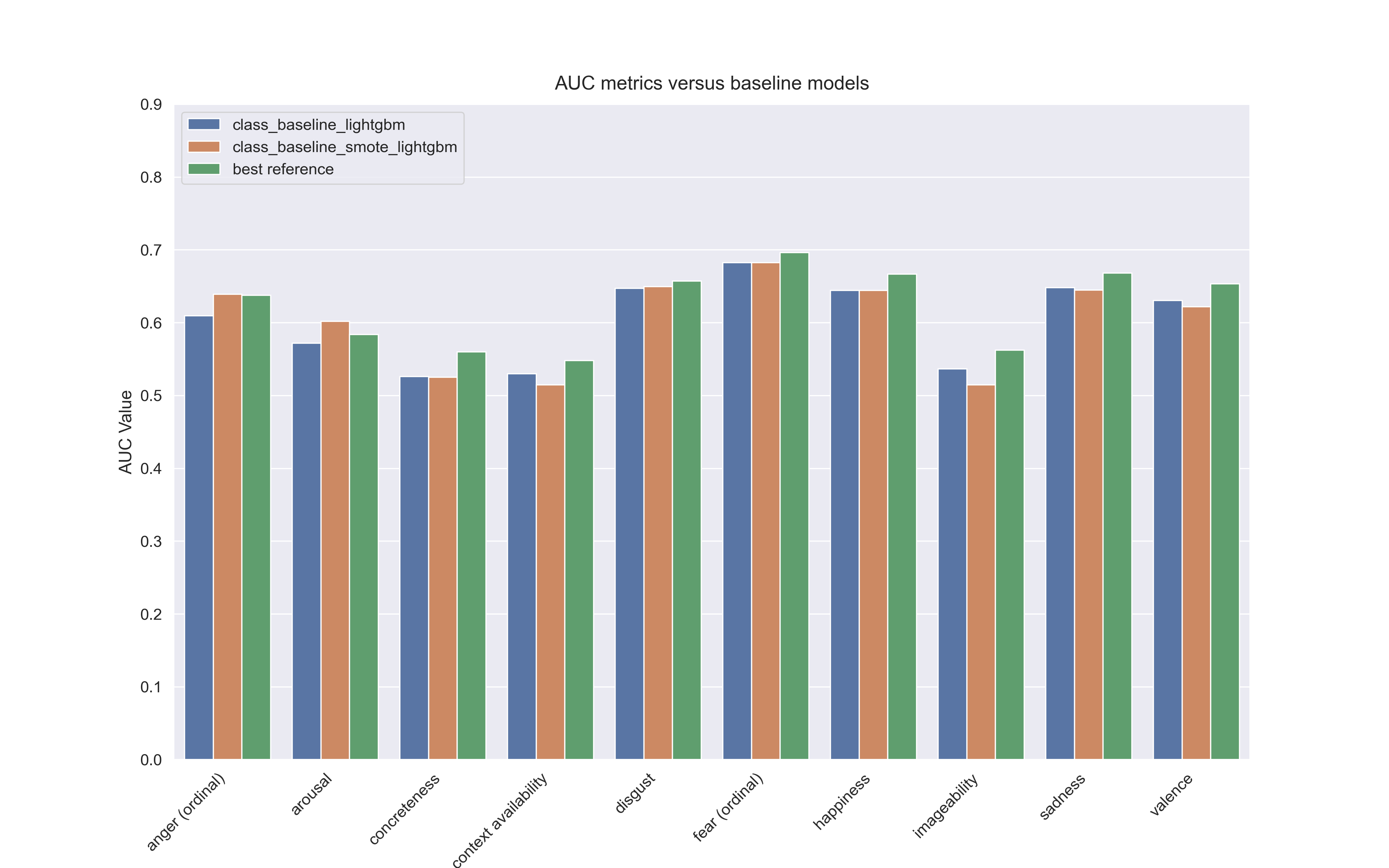}
  \end{tabular} 
  \caption{Comparison of the semi-supervised approach versus baseline models.\label{fig:df_plot_emotions_metrics_vs_baseline}}
\end{figure}

We also include here a comparison of the AUC value with respect to a baseline LigthGBM model (with and without SMOTE) without using semi-supervised learning in Figure \ref{fig:df_plot_emotions_metrics_vs_baseline}. In all the situations where there are significant differences (according to Wilcoxcon signed-rank test), the semi-supervised approach surpasses the baseline models. The only cases where there are no significant differences are "anger (ordinal)" with SMOTE baseline (p-value=0.57), "arousal" (p-value=0.17 for LightGBM, p-value=0.89 for SMOTE variant), "concreteness" (p-value=0.11), "disgust" with SMOTE baseline (p-value=0.34), and "fear (ordinal)" (p-value=0.32). Thus, the semi-supervised approach significantly improves the results for "context availability", "happiness", "sadness" and "valence". The results of the semi-supervised models are also, for all the categories, significantly better than those of a random classifier (AUC=0.5).

\begin{figure}[h!]
\centering
  \begin{tabular}{c@{\qquad}c@{\qquad}c}
  \includegraphics[width=300pt]{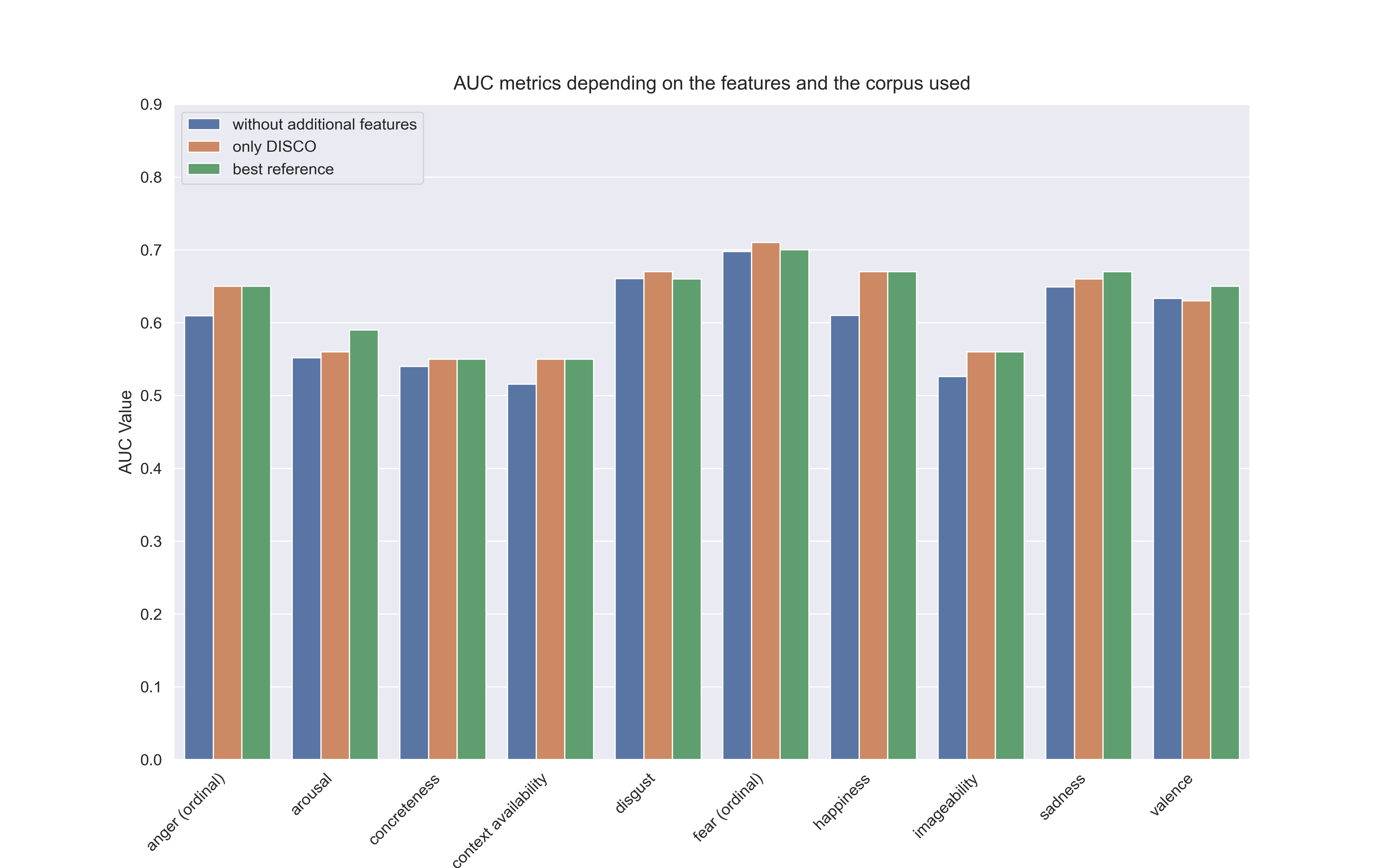}
  \end{tabular} 
  \caption{Comparison of the results without using affective and lexico-semantic additional features, and also when using only DISCO.\label{fig:df_plot_emotions_metrics_q2_q3}}
\end{figure}

Regarding Q2 and Q3, the results are shown in Figure \ref{fig:df_plot_emotions_metrics_q2_q3}. For Q2, there are 7 categories with statistically significant differences according to Wilcoxcon signed-rank test. They are "anger (ordinal)", "arousal", "context availability", "happiness", "imageability", "sadness", and "valence"". In all those cases, the AUC metrics improve when using the affective and lexico-semantic additional features together with the semantic transformers. The AUC improvement in those cases span from 0.02 to 0.06 (with a mean improvement of 0.03).
For Q3, there are only 5 categories where the AUC metrics are statistically significant: "arousal", "disgust", and "valence". Using additional sonnets from the XXth is actually improving the results for "arousal" and "valence" (improvement on AUC of 0.02). For "disgust", using XXth sonnets worsens the AUC metric lowering it on 0.01.

\subsection{Limitations of our Approach}
There are several limitations within our approach. The first one is that the semi-supervised learning training and evaluation uses 270 sonnets from DISCO PAL. It would be interesting to perform it over a bigger corpus of annotated sonnets. However, we think that the corpus size is big enough for carrying out these analyses and obtain statistically meaningful results, since our corpus is has more poems per independent category than other corpora from the literature \parencite{ullrich2017relation}, \parencite{aryani2016measuring}, \parencite{jacobs2017s}, \parencite{obermeier2013aesthetic}, \parencite{haider2020po}. Also, the threshold found using the statistical power serves as another reason to support the statistical analyses carried out.

Another limitation is that the analysis is applied only over highly-structured poems (sonnets) in Castilian. The results may vary if other languages or other types of poems are considered.

\subsection{Software Used}
The main libraries used for the work done in this paper are the following: 
\begin{itemize}
    \item Semi-supervised models \parencite{sklearn-api}
    \item Transformers (\cite{reimers-2019-sentence-bert},  \cite{reimers-2020-multilingual-sentence-bert},  \cite{DBLP:journals/corr/abs-1810-04805}, \cite{CaneteCFP2020})
\end{itemize}

The source code and results for this paper are available in the following repository
\parencite{Barbado2021}.

\section{Conclusion and Future Work}
In this final Section we present some potential future research lines following this work, as well as a summary of the conclusions achieved.

\subsection{Future Work}
In this paper we present a benchmark on supervised modeling of Spanish poetry for inferring the GAM (affective categories), together with lexico-semantic and psychological categories evoked by the poems. Our approach uses semi-supervised learning, since the labelled data set is small. However, there are other alternatives, such as one-shot learning, that could also be used for the supervised modelling task within this context.

It would also be interesting to analysis the results with other types of poems, even combining them within the same corpus with the sonnets. Also, since the semantic transformers are multilingual, and there are available lexicons for lexico-semantic features, the approach could be potentially assessed using poems from other languages. 

Finally, we use as semantic transformers models already pre-trained, but it could be interesting to evaluate the results by fine-tuning the models for poetic texts, since poetry uses linguistic elements, such as metaphors, that may not be properly represented with those models.

\subsection{Conclusions}
This article presents a methodology for inferring the General Affective Meaning (GAM) of Spanish sonnets, along with other lexico-semantic and psychological concepts evoked by the poems. Some of these concepts have binary values, while others are multiclass. We used a labelled subset of sonnets (270) from a bigger sonnet corpus (4572), and we applied a semi-supervised learning approach in order to model the sonnets using as few labelled sonnets as possible, in order to evaluate the results in an statistically significant subset. The sonnets are modelled using sentence transformers, together with lexico-semantic and affective features derived from their individual words by using external lexicons.

According to similar works from the literature, we saw that our proposal achieved good results for at least 76\% of the psychological concepts, and for 60\% of the afective and lexico-semantic ones.

We also checked that the results are significantly better by combining the features derived from the lexicons, together with the sentence transformers, showing how there is information about the sonnets that is not captured by the word embeddings.

Finally, we saw that the prediction of those concepts could be extended for additional sonnets that do not belong to the original corpus used. We included sonnets from other time periods (XXth), and checked how the results are statistically similar.

\subsection{CRediT authorship contribution statement} 
\textbf{Alberto Barbado}: Conceptualization, Investigation, Writing - original draft, Writing - review and editing, Visualization, Formal Analysis, Methodology. 
\textbf{Mar\'{i}a Dolores Gonz\'{a}lez}: Writing - review and editing.
\textbf{D\'{e}bora Carrera}: Writing - review and editing.


\printbibliography


\section{Annex}
\subsection{A.1 - Lexico-semantic and affective features}
\begin{itemize}
    \item \textit{valence\_mean}: mean of valence mean values for the individual words.
    \item \textit{valence\_sd}: mean of valence standard deviation values for the individual words.
    \item \textit{arousal\_mean}: mean of arousal mean values for the individual words.
    \item \textit{arousal\_sd}: mean of arousal standard deviation values for the individual words.
    \item \textit{happiness\_mean}: mean of happiness mean values for the individual words.
    \item \textit{happiness\_sd}: mean of happiness standard deviation values for the individual words.
    \item \textit{anger\_mean}: mean of anger mean values for the individual words.
    \item \textit{anger\_sd}: mean of anger standard deviation values for the individual words.
    \item \textit{sadness\_mean}: mean of sadness mean values for the individual words.
    \item \textit{sadness\_sd}: mean of sadness standard deviation values for the individual words.
    \item \textit{fear\_mean}: mean of fear mean values for the individual words.
    \item \textit{fear\_sd}: mean of fear standard deviation values for the individual words.
    \item \textit{disgust\_mean}: mean of disgust mean values for the individual words.
    \item \textit{disgust\_sd}: mean of disgust standard deviation values for the individual words.
    \item \textit{concreteness\_mean}: mean of concreteness mean values for the individual words.
    \item \textit{concreteness\_sd}: mean of concreteness standard deviation values for the individual words.
    \item \textit{imageability\_mean}: mean of imageability mean values for the individual words.
    \item \textit{imageability\_sd}: mean of imageability standard deviation values for the individual words.
    \item \textit{cont\_ava\_mean}: mean of context availability mean values for the individual words.
    \item \textit{cont\_ava\_sd}: mean of context availability standard deviation values for the individual words.
    \item \textit{max\_arousal}: maximum value of arousal mean values for the individual words.
    \item \textit{min\_arousal}: minimum value of arousal mean values for the individual words.
    \item \textit{max\_valence}: maximum value of valence mean values for the individual words.
    \item \textit{min\_valence}: maximum value of valence mean values for the individual words.
    \item \textit{arousal\_span}: $max\_arousal - min\_arousal$
    \item \textit{valence\_span}: $max\_valence - min\_valence$
    \item \textit{CorAro}: Spearman's correlation between the arousal mean value of the words and their position in the sonnet
    \item \textit{CorVal}: Spearman's correlation between the valence mean value of the words and their position in the sonnet
    \item \textit{AbsCorAro}: absolute value of CorAro
    \item \textit{AbsCorVal}: absolute value of CorVal
    \item \textit{sigma\_aro}: $\frac{arousal\_mean}{1/\sqrt{N}}$ with N the number of words in the sonnet.
    \item \textit{sigma\_val}: $\frac{valence\_mean}{1/\sqrt{N}}$ with N the number of words in the sonnet.
\end{itemize}

\end{document}